\documentclass[conference]{IEEEtran}

\usepackage[cmex10]{amsmath}
\usepackage{amsthm}
\usepackage{amssymb}
\usepackage{mathrsfs}
\usepackage{graphicx}
\usepackage{float}
\usepackage{array}
\usepackage{epstopdf}
\usepackage{multirow}
\usepackage{cite}
\usepackage{xcolor}
\usepackage[export]{adjustbox}

\begin{document}

\title{Age and Gender Prediction From Face Images Using Attentional Convolutional Network}

\author{Amirali Abdolrashidi$^1$, Mehdi Minaei$^2$, Elham Azimi$^3$, Shervin Minaee$^4$\\
$^1$University of California, Riverside\\
$^2$Sama Technical College, Azad University\\ 
$^3$New York University\\ 
$^4$Snap Inc \\ \\
}

\maketitle

\begin{abstract}
Automatic prediction of age and gender from face images has drawn a lot of attention recently, due it is wide applications in various facial analysis problems.
However, due to the large intra-class variation of face images (such as variation in lighting, pose, scale, occlusion), the existing models are still  behind the desired accuracy level, which is necessary for the use of these models in real-world applications. 
In this work, we propose a deep learning framework, based on the ensemble of attentional and residual convolutional networks, to predict gender and age group of facial images with high accuracy rate. 
Using attention mechanism enables our model to focus on the important and informative parts of the face, which can help it to make a more accurate prediction.
We train our model in a multi-task learning fashion, and augment the feature embedding of the age classifier, with the predicted gender, and show that doing so can further increase the accuracy of age prediction.
Our model is trained on a popular face age and gender dataset, and achieved promising results.
Through visualization of the attention maps of the train model, we show that our model has learned to become sensitive to the right regions of the face.
\end{abstract}

\IEEEpeerreviewmaketitle
%Revise

\section{Introduction}
\label{sec:Intro}
Age and gender information are very important for various real world applications, such as social understanding, biometrics, identity verification, video surveillance, human-computer interaction, electronic customer, crowd behavior analysis, online advertisement, item recommendation, and many more.
Despite their huge applications, being able to automatically predicting age and gender from face images is a very hard problem, mainly due to the various sources of intra-class variations on the facial images of people, which makes the use of these models in real world applications limited.

There are numerous works proposed for age and gender prediction in the past several years.
The earlier works were mainly based on hand-crafted features extracted facial images followed by a classifier.
But with the great success of deep learning models in various computer vision problems in the past decade \cite{krizhevsky2012imagenet, ren2015faster, minaee2019biometric, chen2017deeplab, minaee2019mtbi}, the more recent works on age and gender predictions are mostly shifted toward deep neural networks based models.

\begin{figure}[t]
\begin{center}
   \includegraphics[page=10, width=0.64\linewidth]{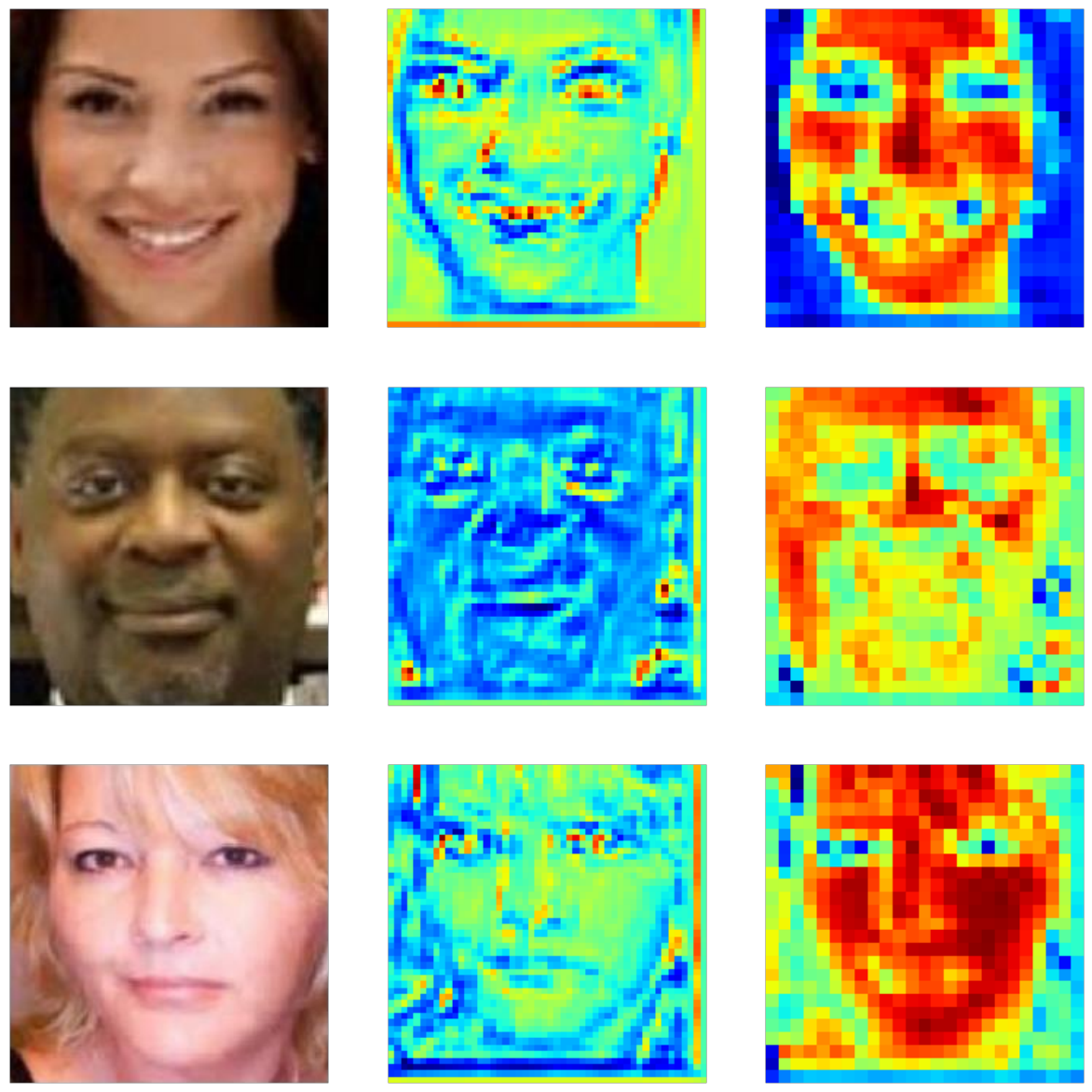}
\end{center}
   \caption{Three sample face images, and their corresponding attention map outputs, from two different layers. Color scale is from blue (lowest) to red (highest).
   }
\label{fig:att_map_3}
\end{figure}

In this work, we propose a deep learning framework to jointly predict the age and gender from face images. 
Given the intuition that some local regions of the face have more clear signals about the age and gender of an individual (such as beard and mustache for male, and wrinkles around eyes and mouth for age), we use an attentional convolutional network as one of our backbone models, to better attend to the salient and informative part of the face. 
Figure \ref{fig:att_map_3} provide three sample images, and the corresponding attention map outputs of two different layers of our model for these images.
As we can see, the model outputs are mostly sensitive to the edge patterns around facial parts, as well as wrinkles, which are important for age and gender prediction.

As predicting age and gender from faces are very related, we use a single model with multi-task learning approach to jointly predict both gender and age bucket.
Also, given that knowing the gender of someone, we can better estimate her/his age, we augment the feature of the age-prediction branch with the predicted gender output.
Through experimental results, we show that adding the predicted gender information to the age-prediction branch, improves the model performance.
To further improve the prediction accuracy of our model, we combine the prediction of attentional network with the residual network, and use their ensemble model as the final predictor.

Here are the contributions of this work:
\begin{itemize}
    \item We propose a multi-task learning framework to jointly predict the age and gender of individuals from their face images. 
    \item We develop an ensemble of attentional and residual networks, which outperforms both individual models. The attention layers of our model learn to focus on the most important and salient parts of the face.
    \item We further propose to feed the predicted gender label to the age prediction branch, and show that doing this will improve the accuracy of age prediction branch.
    \item With the help of the attention mechanism, we can explain the predictions of the classifiers after they are trained, by locating the salient facial regions they are focusing on each image.
\end{itemize}

The structure of the remaining parts of this paper is as follows.
Section \ref{sec:related} provides an overview of some of the previous works on age and gender prediction.
Section \ref{sec:Framework} provides the details of our proposed framework, and the architecture of our multi-task learning model.
Section \ref{sec:dataset}, provides a quick overview of the dataset used in our framework.
Then, in Section \ref{sec:Evaluation}, we provide the experimental studies, the quantitative performance of our model, and also visual evaluation of model outputs.
Finally the paper is concluded in Section \ref{sec:Conclusion}.

%% https://www.cv-foundation.org/openaccess/content_cvpr_workshops_2015/W08/papers/Levi_Age_and_Gender_2015_CVPR_paper.pdf
\section{Related Works}
\label{sec:related}
Face is one of the most popular biometrics (along with fingerprint, iris, and palmprint \cite{zhao2009direct, minaee2015fingerprint, de2016iris, minaee2016experimental, minaee2015highly, mistani2011multispectral}), and face recognition and facial attributes/characteristics prediction have attracted a lot of attention in the past few decades \cite{wright2008robust, parkhi2015deep, minaee2017face}.
Age and gender prediction from face images, as a specific face analysis problem have also drawn attention in the recent years, and there have been several previous works for age and gender prediction from face images. 
Here we provide an overview of some of the most promising works.

\begin{figure}[h]
\begin{center}
   \includegraphics[page=3,width=0.99\linewidth]{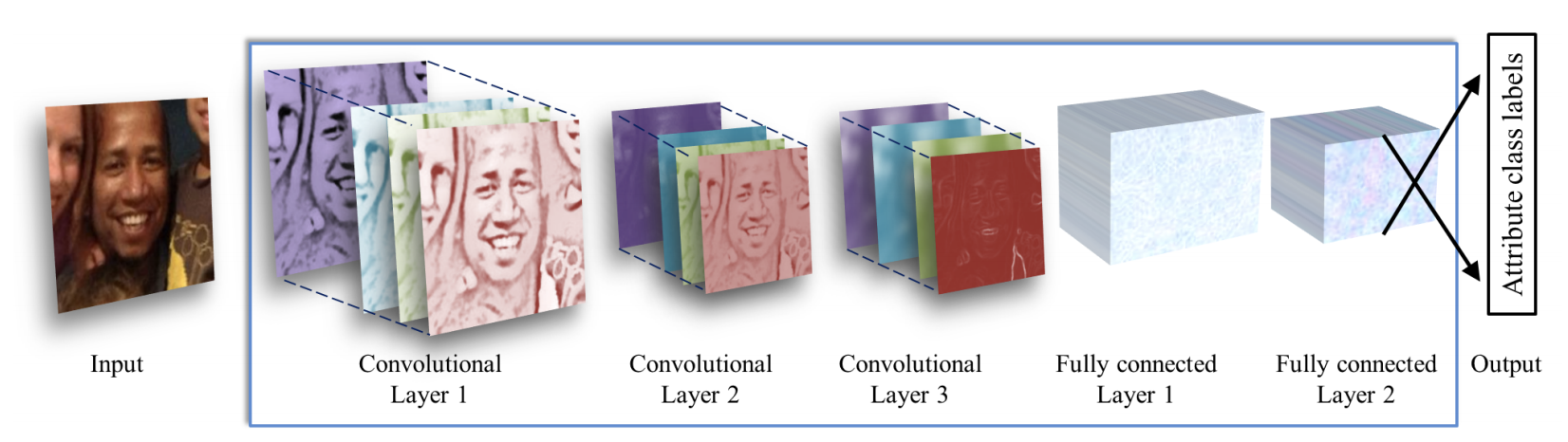}
\end{center}
   \caption{The architecture of the  work by Levi and Hassner, courtesy of \cite{levi2015age}.}
\label{fig:levi}
\end{figure}

% Age and Gender Classification using Convolutional Neural Networks
In \cite{levi2015age}, Levi and Hassner proposed a simple convolutional neutral network architecture with 5 layers, to perform age and gender prediction. 
Despite the simplicity of this model, they achieved promising results on Adience benchmark for age and gender estimation. 
Figure \ref{fig:levi} shows the architecture of the model proposed in \cite{levi2015age}.

% A hybrid deep learning CNN–ELM for age and gender classification
In \cite{duan2018hybrid}, Duan et al. proposed a hybrid structure, which combines Convolutional Neural Network (CNN) and Extreme Learning Machine (ELM), to perform age and gender classification.
CNN is used to extract the features from the input images, while ELM classifies the intermediate results.
They were able to obtain reasonable accuracy on the MORPHII and Adience Benchmarks.

% How transferable are CNN-based features for age and gender classification?
In \cite{ozbulak2016transferable}, Ozbulak et al. analyzed the transferability of existing deep convolutional neural network (CNN) models for age and gender classification. The generic AlexNet-like architecture
and domain specific VGG-Face CNN model are fine-tuned with the Adience dataset prepared for age and gender classification in uncontrolled environments. 
Not surprisingly, they were able to obtain promising results on the features learned from these popular architectures.

\begin{figure}[h]
\begin{center}
   \includegraphics[page=3,width=0.85\linewidth]{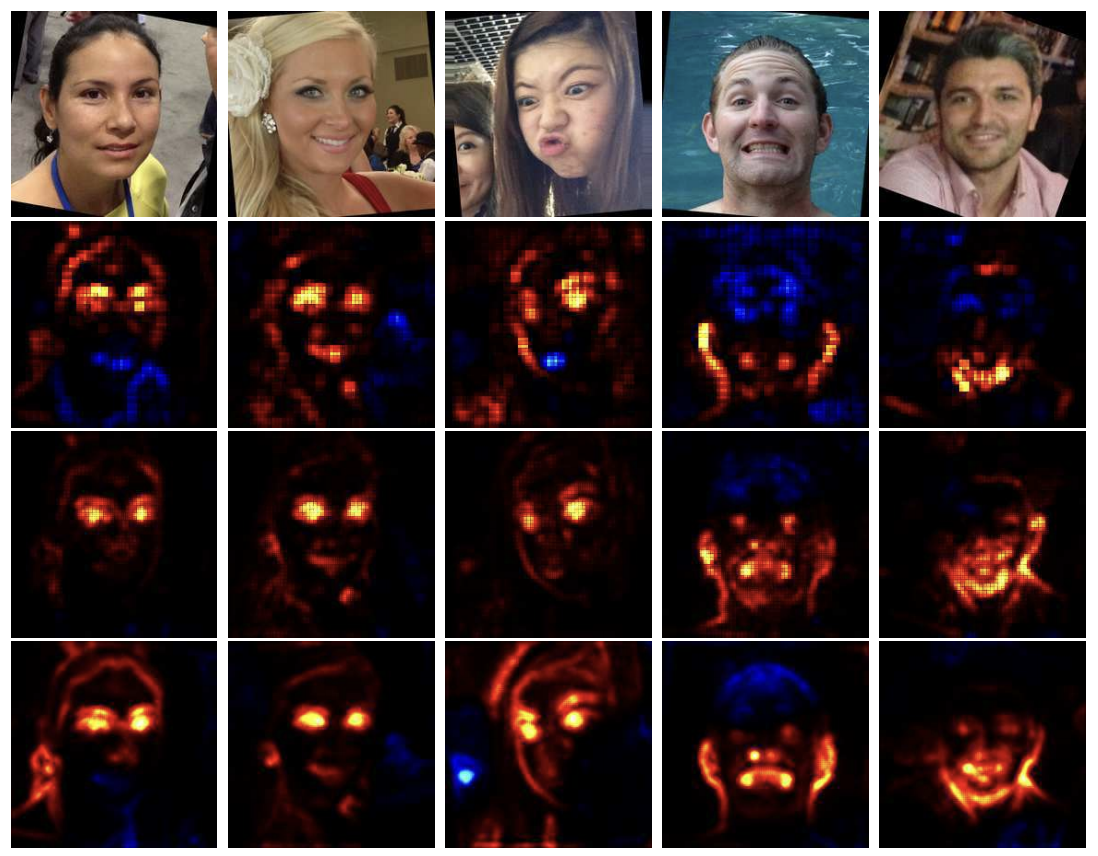}
\end{center}
   \caption{From top to bottom: Input image, followed by relevance
maps for the best performing CaffeNet, GoogleNet and the VGG16 model for gender prediction. 
Courtesy of \cite{lapuschkin2017understanding}.}
\label{fig:rel_map}
\end{figure}

% Understanding and comparing deep neural networks for age and gender classification
In \cite{lapuschkin2017understanding}, Lapuschkin et al.  compared four popular neural network architectures, studied the effect of pre-training, evaluated the robustness of the considered alignment preprocessings via cross-method test set swapping, and intuitively visualized the model’s prediction strategies in given pre-processing conditions using the Layer-wise Relevance Propagation (LRP) algorithm.
They were able to obtain very interesting relevance maps for some of the popular model architectures, as shown in Figure \ref{fig:rel_map}. 
\cite{levi2015age}.

% Adversarial spatial frequency domain critic learning for age and gender classification

\begin{figure}[h]
\begin{center}
   \includegraphics[page=3,width=0.99\linewidth]{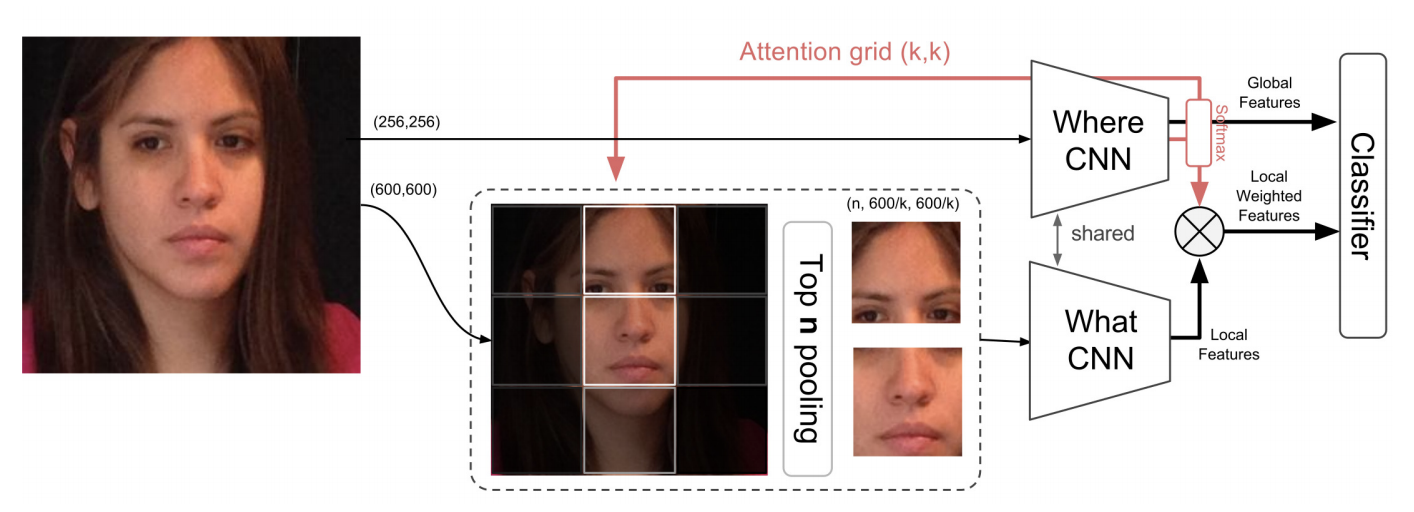}
\end{center}
   \caption{The proposed attention model by Rodriguez et al. 
Courtesy of \cite{rodriguez2017age}.}
\label{fig:ffatt}
\end{figure}

% Age and gender recognition in the wild with deep attention
In \cite{rodriguez2017age}, Rodriguez et al. proposed a novel feedforward attention mechanism that is able to discover the most informative and reliable parts of a given face for improving age and gender classification. 
More specifically, given a downsampled facial image, the proposed model is trained based on an end-to-end learning framework to extract the most discriminative patches from the original high-resolution image.
They were able to obtain promising results on several benchmarks, including  Adience, Images of Groups, and MORPH II.
The block-diagram of this work is shown in Figure \ref{fig:ffatt}.

Some of the other promising works for age and gender predictions includes: adversarial spatial frequency domain critic learning \cite{lee2018adversarial}, Region-SIFT and multi-layered SVM \cite{kim2018age}, and Landmark-Guided Local Deep Neural Networks \cite{zhang2018landmark}.

\begin{figure*}[t]
\begin{center}
   \includegraphics[width=0.9\linewidth]{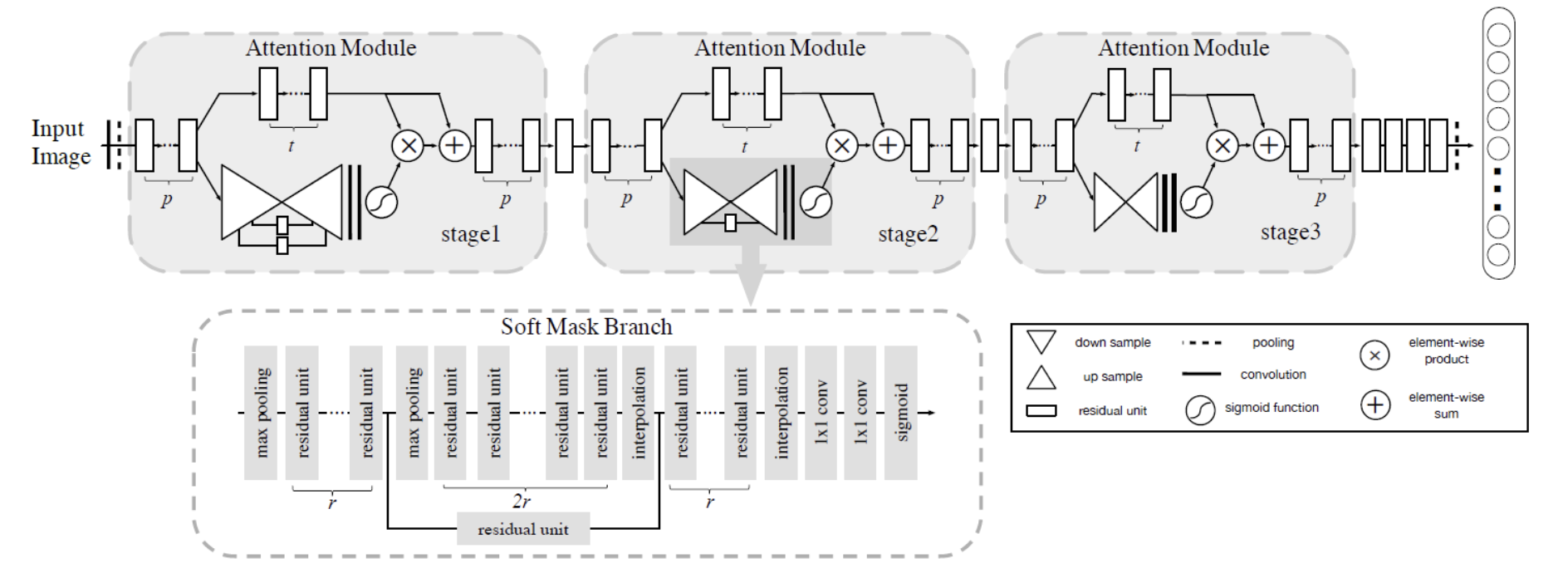}
\end{center}
   \caption{The architecture of residual attention network, courtesy of \cite{wang2017residual}.}
\label{fig:RAN}
\end{figure*}

\section{The Proposed Framework}
\label{sec:Framework}
\begin{figure}[h]
   \includegraphics[width=0.82\linewidth, left]{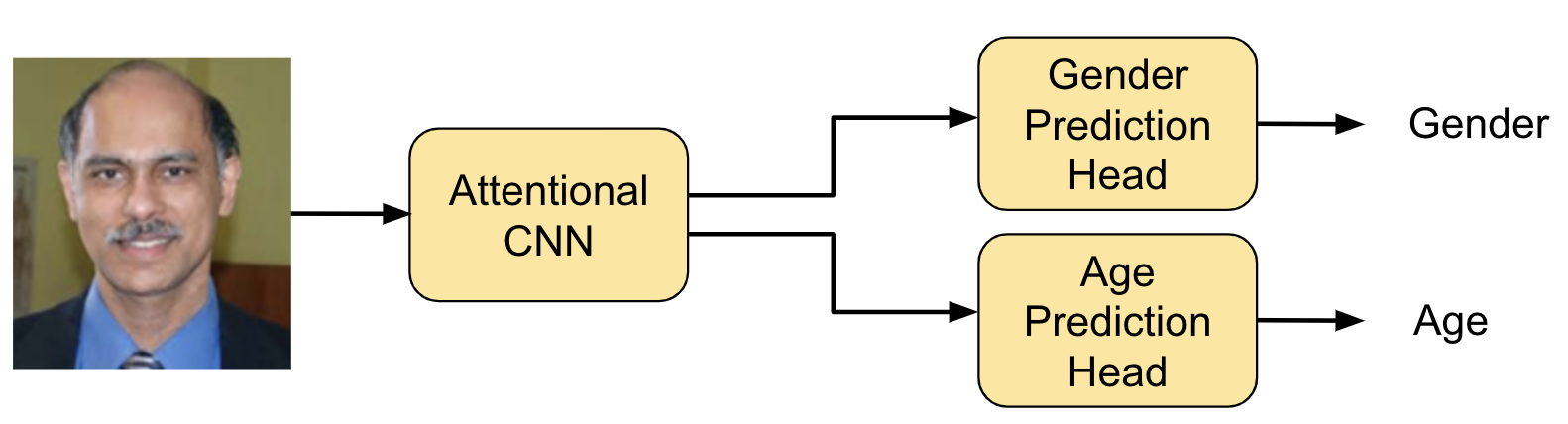}
   \caption{The block-diagram of a multi-task learning network for joint age and gender prediction.}
\label{fig:multitask}
\end{figure}

In this section we provide the details of the proposed age and gender prediction framework.
We formulate this as a multi-task learning problem, in which a single model is used to predict both gender and age-buckets simultaneously. 
In another word, a single convolutional neural network with two heads (output branches) is used to jointly predict age and gender.
Figure \ref{fig:multitask} shows the block-diagram of a simple multi-task learning model for joint age and gender prediction.

\begin{figure}[h]
   \includegraphics[width=0.99\linewidth,left]{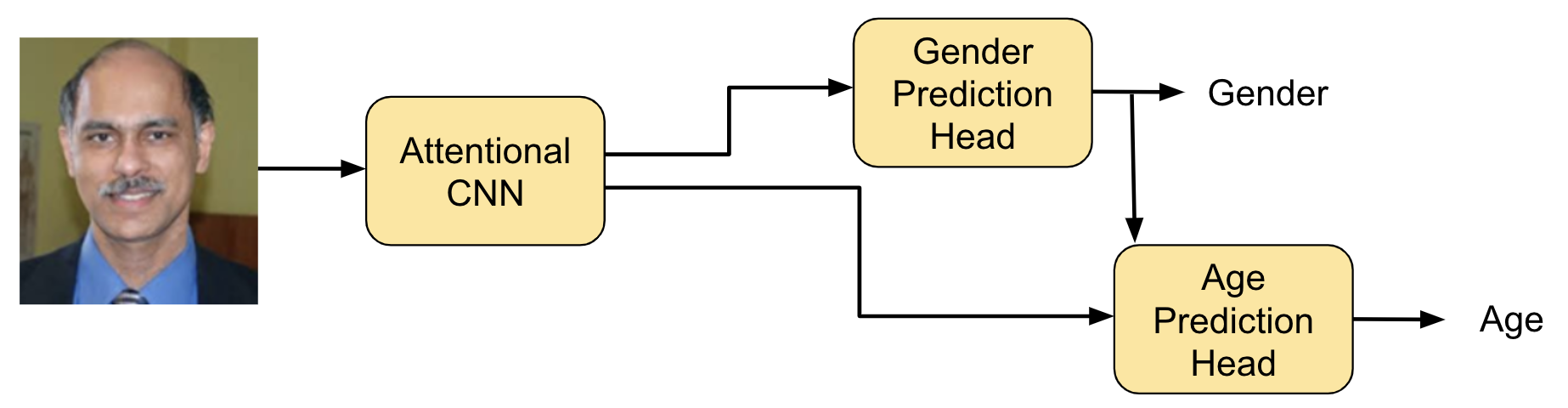}
   \caption{The block-diagram of the framework for joint age and gender prediction, in which the predicted gender is used along with the neural embedding as the input for the age prediction branch.}
\label{fig:gender_augment}
\end{figure}

Given the intuition that knowing one's gender, can enable us to better predict her/his age, we augment the input feature of the age prediction part of the model, with the the predicted-gender from the other head. 
This can help us the age model to have access to a rough estimation of gender.
Through experimental study, we show that doing so improves the performance of the age prediction.
Figure \ref{fig:gender_augment}, provides the block diagram of the proposed model architecture.

To further boost the accuracy of our model, we propose to use an ensemble of attentional network, and a residual convolutional network. 
Once these models are trianed, their average predictions (output probabilities) are used as the final prediction.
We will give more details on the architecture of each of these two models in the below parts.

\subsection{Residual Attentional Network}
As mentioned above, an important piece of our framework is the residual attentional network (RAN), which is a convolutional neural network with attention mechanism, which can be incorporated with state-of-art feed forward network architecture in an end-to-end training fashion \cite{wang2017residual}.
This network is built by stacking attention modules, which generate attention-aware features that adaptively changes as layers go deeper into the network.

The composition of the Attention Module includes two branches: the trunk branch and the mask branch.
Trunk Branch performs feature processing with Residual Units.
Mask Branch uses bottom-up top-down structure softly weight output features with the goal of improving trunk branch features.
Bottom-Up Step: collects global information of the whole image by downsampling (i.e. max pooling) the image.
Top-Down Step: combines global information with original feature maps by upsampling (i.e. interpolation) to keep the output size the same as the input feature map.
The full architecture of residual attention network is shown in Figure \ref{fig:RAN}.

\subsection{ResNet Model}
\begin{figure}[htb]
\begin{center}
   \includegraphics[page=2,width=0.95\linewidth]{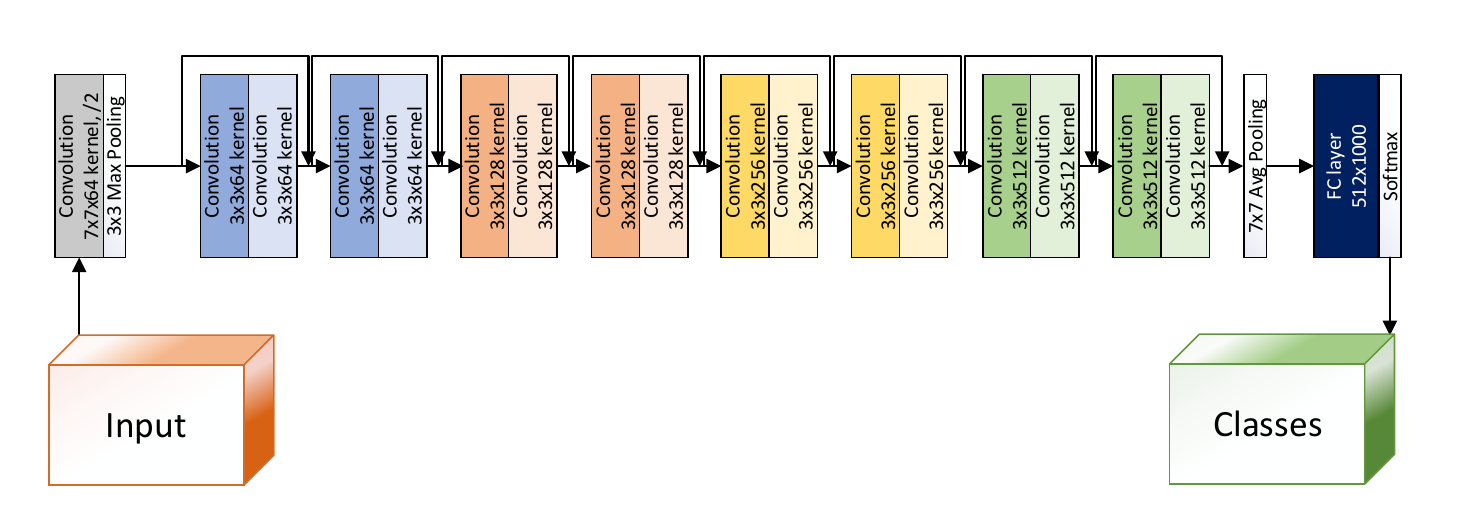}
\end{center}
   \caption{The architecture of ResNet18 model. Courtesy of \cite{he2016deep}.}
\label{fig:ResnetNN}
\end{figure}

Another model used in our framework is based on residual convolutional network (ResNet) \cite{he2016deep}. 
ResNet is known to have a better gradient flow by providing the skip connection in each residual block.
Here we use a ResNet architecture to perform gender classification on the input image. 
Then, the predicted output of the gender branch is concatenated with the last hidden layer of the age branch.
The overall block diagram of ResNet18 model is illustrated in Figure \ref{fig:ResnetNN}.
ResNet50 architecture is pretty similar to ResNet18, the main difference being having more layers.

\subsection{Ensemble of Attentional and Residual Networks}
To boost the prediction accuracy of our model, we use the ensemble of the above two architectures, and combine their predictions. 
Through experimental results, we show that the ensemble model achieves higher performance than both of the individual models.

\section{UTK-Face Dataset}
\label{sec:dataset}
\begin{figure}[htb]
\begin{center}
   \includegraphics[page=11,width=0.99\linewidth]{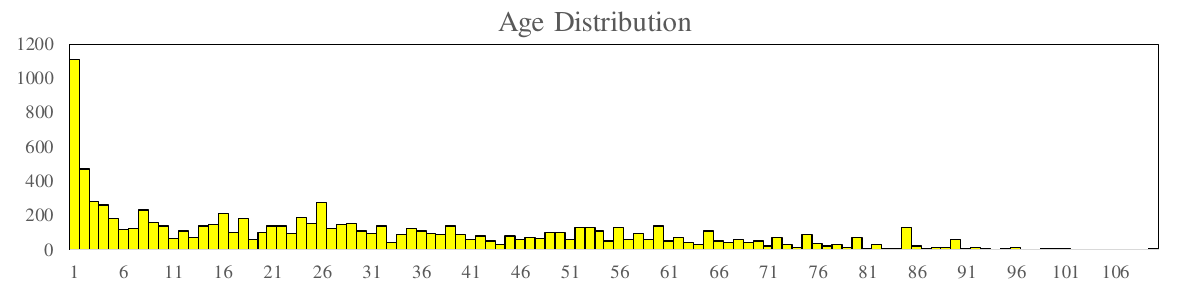}
\end{center}
   \caption{Age distribution in the used UTK-Face dataset.}
\label{fig:utk_age_dist}
\end{figure}

In this section, we provide a quick overview of the dataset used in this work, UTK-Face dataset \cite{utk}. We used the cropped and aligned version of this dataset, which consists of 9780 images of size 200x200 (4372 male, 5408 female). The images are taken from the faces of people from various ethnic groups and ages (1 to 110 years). The age distribution of the used dataset is shown in Figure \ref{fig:utk_age_dist}. It can be seen that the largest portion of the images belong to people under 2 years of age.

\begin{figure}[htb]
\begin{center}
   \includegraphics[page=1,width=0.9\linewidth]{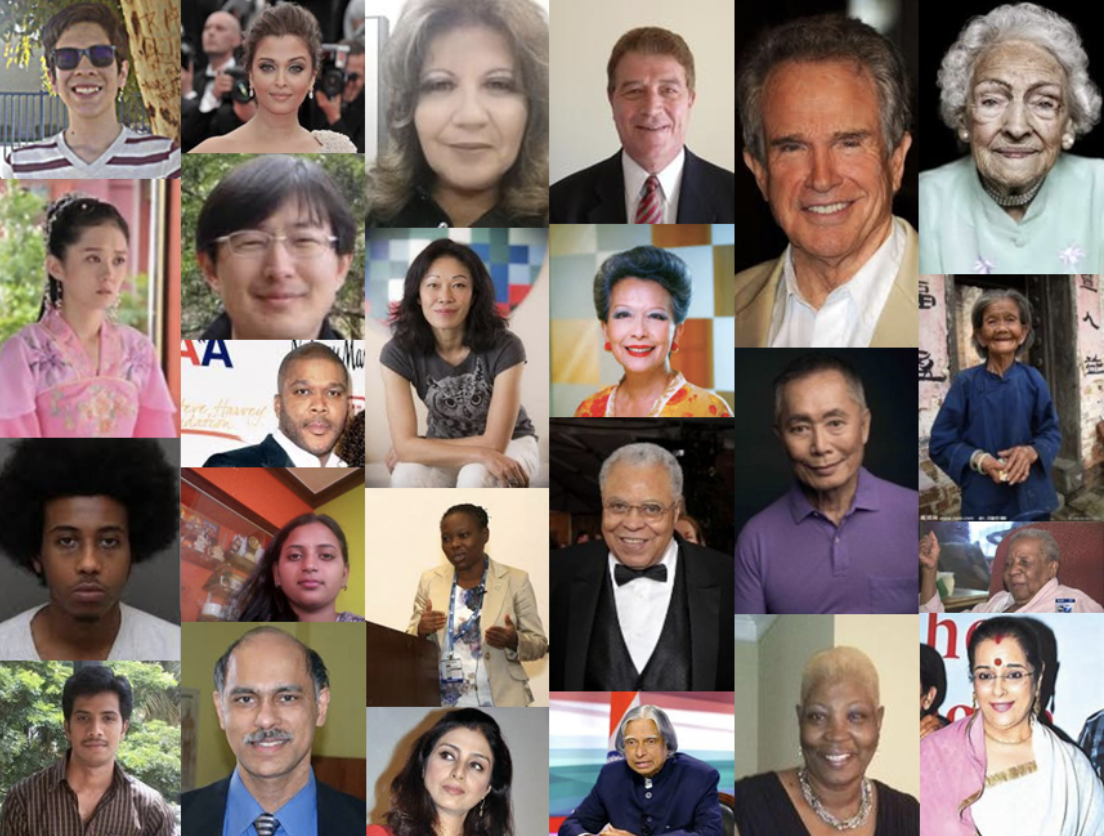}
\end{center}
   \caption{Sample images from UTK-Face database.}
\label{fig:utk_face_samples}
\end{figure}

Also, Figure \ref{fig:utk_face_samples} denotes several images from UTK-Face dataset. 
As we can see frm this figure, UKT-Face dataset has a good diversity in terms of age and gender.
It is worth to note that, in order to make the age prediction simpler, we group people whose age is in some range into the same age-bucket (such as 0-10), and try to predict their age-bucket instead of the exact age. Doing so, on one hand makes age prediction simpler, and on the other hand turns this model from regression to classification (which does not need any clipping of the predicted values).

\section{Experimental Results}
\label{sec:Evaluation}
Before presenting the quantitative and qualitative performance of the proposed framework, let us first discuss about the hyper-parameter values used in our training procedure.
These values are tuned based on the model performance on the validation set.

\subsection{Model Hyper-parameter Values}
The proposed model has been trained for over 100 epochs on an Nvidia Tesla GPU.
The batch size is set to 16, and number of workers to 8. 
ADAM optimizer with a learning rate of 0.005 is used to optimize the loss function.
PyTorch library is used for the implementation of experimental studies \cite{paszke2017automatic}.

\subsection{Model Quantitative Performance}
In this part, we provide the model performance in terms of various metrics, such as classification accuracy for both branches (gender prediction, and age-bucket classification), as well as "average age-bucket absolute difference (AABD)".
AABD calculates the absolute difference between the ground-truth and the predicted age-bucket of each person, and average them out among all test samples. 
As an example, if someone's age is 25 (its age-bucket being 20-30, therefore 3), and the predicted age by model is 4, then the AABD value will be 1. 

As mentioned earlier, each model predicts a probability vector, which shows the likelihood of that image being in each of the possible classes (for example, male vs female, for gender model). 
In the ensemble model, the predicted probabilities from Attentional-CNN and ResNet are averaged and used to infer the predicted classes of each task. 
In Table \ref{tab:accuracy_results}, we presents the comparison between the proposed ensemble model, with each of the two individual models. 
As we can see the ensemble model outperforms both the individual models on both tasks (gender and age prediction).
In Table \ref{tab:accuracy_results2}, we compare the performance of the proposed model with one of the previous works, on gender classification task.

% \begin{figure}[htb]
% \begin{center}
%   \includegraphics[page=6,width=0.7\linewidth]{img/Figures.pdf}
% \end{center}
%   \caption{Accuracy results}
% \label{fig:accuracy_results}
% \end{figure}

\begin{table}[htb]
    \centering
    \caption{Comparison of accuracy of different models}
    \begin{tabular}{|l|c|c|c|}
        \hline
         Model & Age Range Acc & Gender Acc & Age Bucket Diff
         \\ \hline 
         Attention CNN & 0.742 & 0.552 & 0.33
         \\ \hline
         ResNet & 0.900 & 0.965 & 0.14
         \\ \hline
         The Ensemble Model & 0.913 & 0.965 & 0.11
         \\ \hline
    \end{tabular}
    \label{tab:accuracy_results}
\end{table}

\begin{table}[htb]
    \centering
    \caption{Comparison with a previous work.}
    \begin{tabular}{|l|c|c|c|}
        \hline
         Model & Gender Acc
         \\ \hline
         %\cite{levi2015age} (Not UTKFace) & 0.9094
         %\\ \hline
         %\cite{al2019human} (Not UTKFace) & 0.9134
         %\\ \hline
         Automatic Gender Detection \cite{nair2019automated} & 0.9485
         \\ \hline
         Our model & 0.965
         \\ \hline
    \end{tabular}
    \label{tab:accuracy_results2}
\end{table}

\subsection{Model Predicted Scores}
\begin{figure}[htb]
\begin{center}
   \includegraphics[page=12,width=0.9\linewidth]{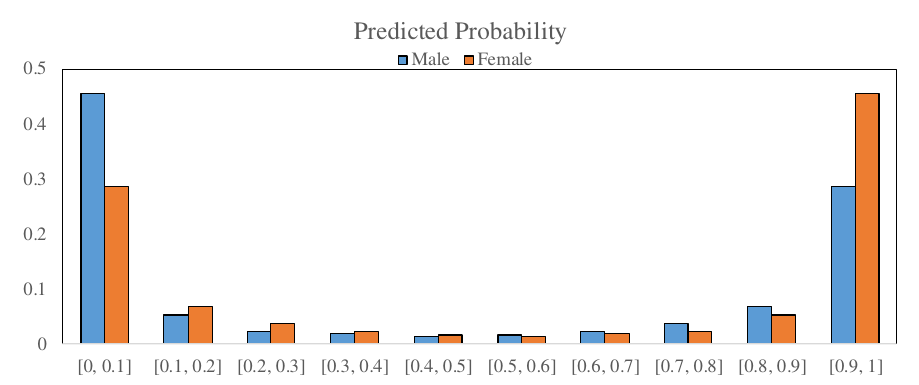}
\end{center}
   \caption{Predicted probability distribution for gender classification}
\label{fig:pp_gender}
\end{figure}

Figure \ref{fig:pp_gender} shows the histogram of the predicted probability scores for gender by our model, for each class. 
Note that here we are showing the histogram of the likelihood of an image being female. 
It can be seen that in the majority of the cases, the model's prediction of gender is much higher in the extremes, i.e. the model makes its decision with remarkable certainty. On the other hand, more uncertain decisions, which are shown in the middle of the chart, are much lower in frequency.

\subsection{Model Predictions for Some Sample Images}
\begin{figure}[htb]
\begin{center}
   \includegraphics[page=9,width=0.99\linewidth]{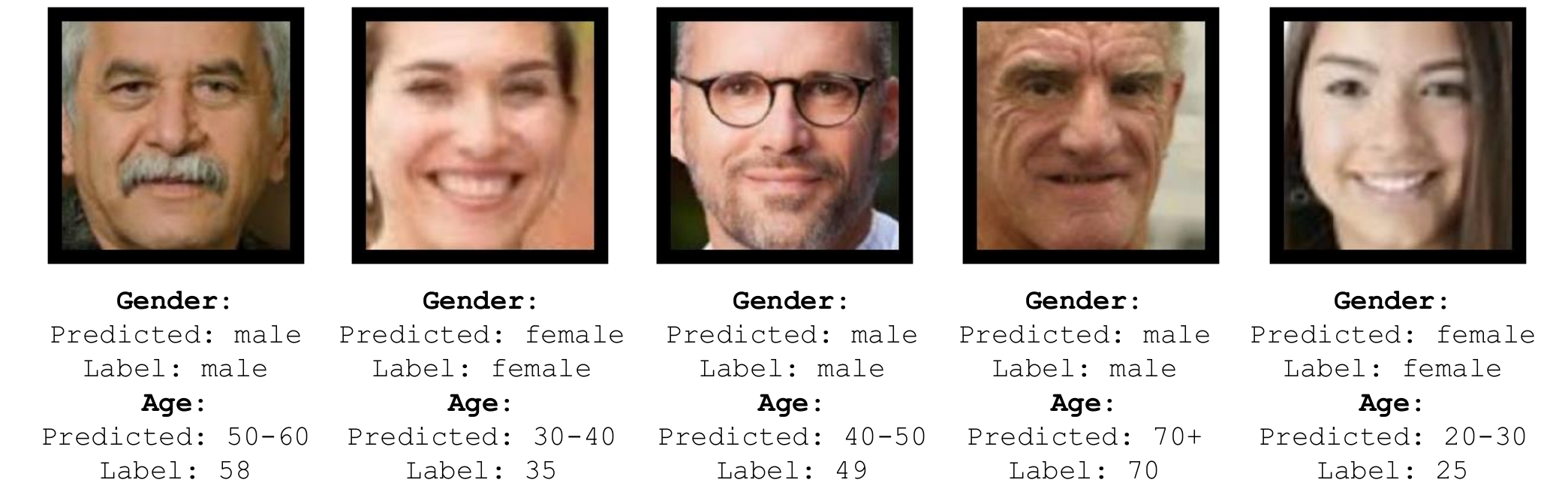}
\end{center}
   \caption{Classification results for sample images}
\label{fig:sample_results}
\end{figure}

We also present the model prediction for some of the sample images of our test set.
In Figure \ref{fig:sample_results}, some of the model's predictions are provided next to the true labels of the face image.
It can be seen that, for most of these sample images, the true age label falls within the predicted age range by our model.

\subsection{Model Confusion Matrix}
\begin{figure}[htb]
\begin{center}
   \includegraphics[page=4,width=0.9\linewidth]{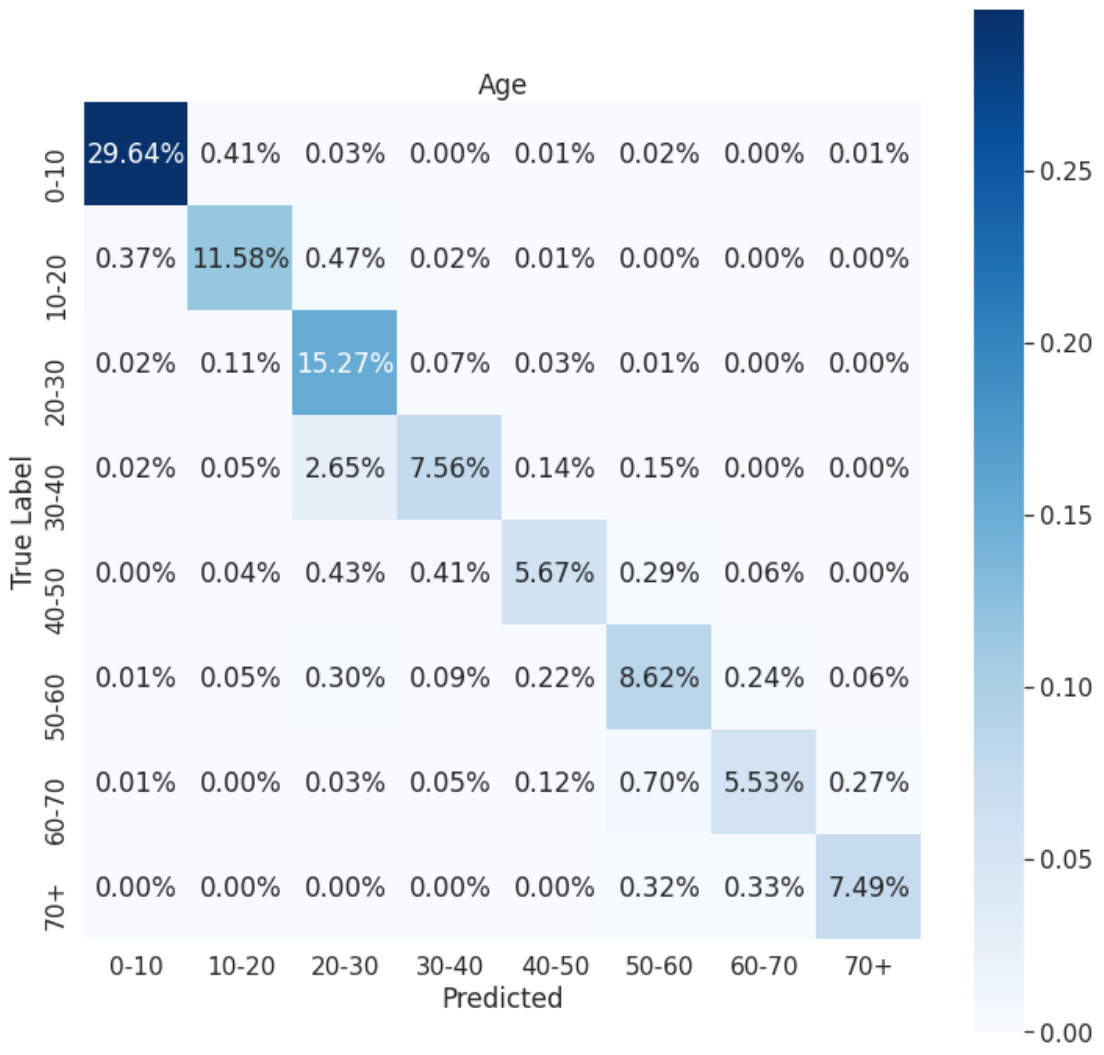}
\end{center}
   \caption{Confusion matrix for age range}
\label{fig:confusion_age}
\end{figure}

\begin{figure}[htb]
\begin{center}
   \includegraphics[page=5,width=0.5\linewidth]{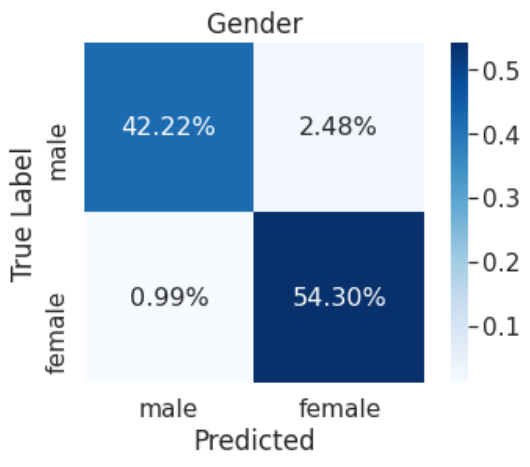}
\end{center}
   \caption{Confusion matrix for gender outputs}
\label{fig:confusion_gender}
\end{figure}

To provide a more detailed analysis of the proposed model's performance, we also present the confusion matrix of this model for each of the two tasks.
Figure \ref{fig:confusion_age} and Figure \ref{fig:confusion_gender} show the confusion matrix for age range and gender classification respectively. It can clearly be seen that a vast majority of the cases are predicted as the true label of their class, seen on the main diagonal of the matrix. 
It is worth noting the largest false positive percentage in the age range classification model, corresponds to the images belonging to the 30-40 age range which are mistaken for 20-30. %(which could be justified as people in the 25-35 years old range, may look quite similar).

\subsection{Model Attention Map Visualization}
\begin{figure*}
\begin{center}
   \includegraphics[page=1,width=0.8\linewidth]{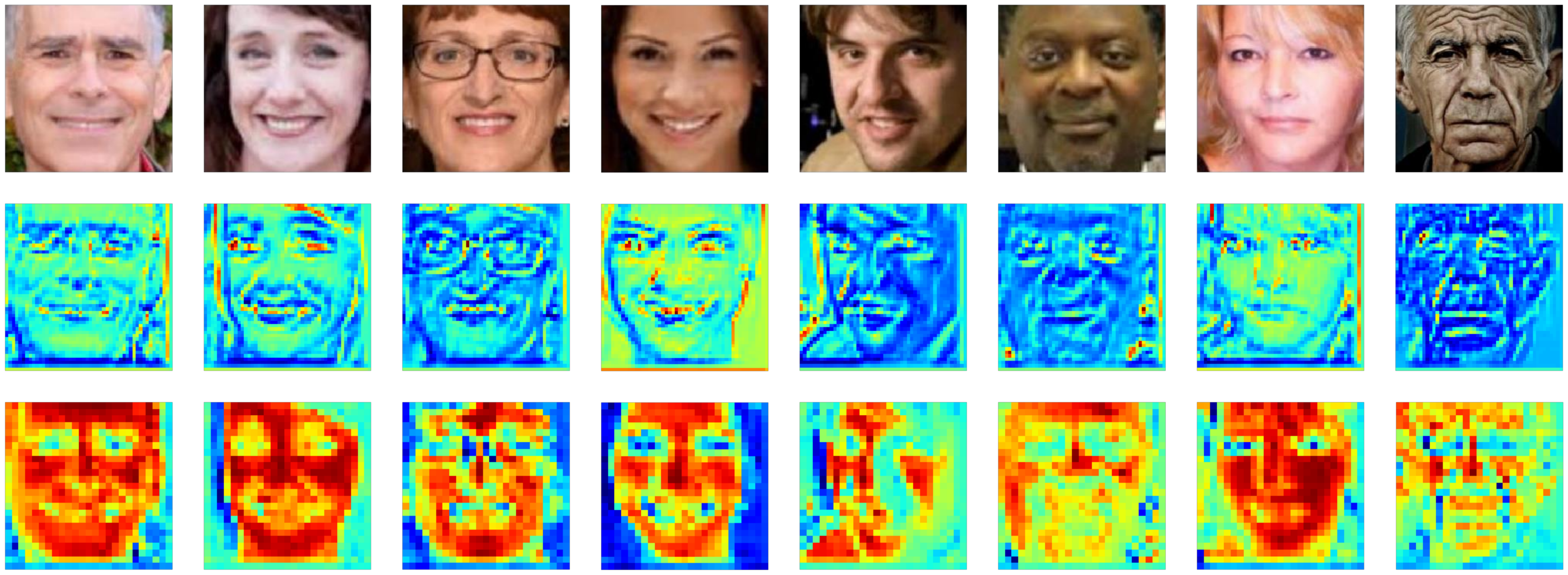}
\end{center}
   \caption{The attention maps of our model, for eight sample images.}
\label{fig:utk_with_attention}
\end{figure*}

In Figure \ref{fig:utk_with_attention}, we show the output of two of the attention maps (from the three attention modules in our model) of the trained model for a few sample images from the test set. 
As it can be seen from this figure, our model is learned to extract features from different salient parts of the face, such as the outline, eyes, and wrinkles, which sounds reasonable when trying to make prediction about someone's age and gender.

\section{Conclusion}
\label{sec:Conclusion}
In this work we propose a multi-task learning framework, to simultaneously predict age and gender from face images. Our framework is based on an ensemble of ResNet-based model and an attention-based model.
We trained and tested the proposed model on  the UTKFace dataset consisting a large variety of faces from different ages, genders and ethnicities. 
Through experimental studies, we show that the prediction accuracy of the ensemble model (for both age and gender prediction tasks) surpasses in those of the separated models.
We also showed that providing the prediction of the gender model as one of the input signal for the age-prediction branch, can improve the accuracy of predicted age values.
Through visualization of the attention maps of the trained model, we show that the model learned to focus on the most salient part of the face, useful for predicting age and gender.

% \section*{Acknowledgment}
% We would like to thank ...

%\begin{thebibliography}{1}
%\end{thebibliography}

\bibliographystyle{IEEEtran}
%\bibliography{ref}

\begin{thebibliography}{10}
\providecommand{\url}[1]{#1}
\csname url@samestyle\endcsname
\providecommand{\newblock}{\relax}
\providecommand{\bibinfo}[2]{#2}
\providecommand{\BIBentrySTDinterwordspacing}{\spaceskip=0pt\relax}
\providecommand{\BIBentryALTinterwordstretchfactor}{4}
\providecommand{\BIBentryALTinterwordspacing}{\spaceskip=\fontdimen2\font plus
\BIBentryALTinterwordstretchfactor\fontdimen3\font minus
  \fontdimen4\font\relax}
\providecommand{\BIBforeignlanguage}[2]{{%
\expandafter\ifx\csname l@#1\endcsname\relax
\typeout{** WARNING: IEEEtran.bst: No hyphenation pattern has been}%
\typeout{** loaded for the language `#1'. Using the pattern for}%
\typeout{** the default language instead.}%
\else
\language=\csname l@#1\endcsname
\fi
#2}}
\providecommand{\BIBdecl}{\relax}
\BIBdecl

\bibitem{krizhevsky2012imagenet}
A.~Krizhevsky, I.~Sutskever, and G.~E. Hinton, ``Imagenet classification with
  deep convolutional neural networks,'' in \emph{Advances in neural information
  processing systems}, 2012, pp. 1097--1105.

\bibitem{ren2015faster}
S.~Ren, K.~He, R.~Girshick, and J.~Sun, ``Faster r-cnn: Towards real-time
  object detection with region proposal networks,'' in \emph{Advances in neural
  information processing systems}, 2015, pp. 91--99.

\bibitem{minaee2019biometric}
S.~Minaee, A.~Abdolrashidi, H.~Su, M.~Bennamoun, and D.~Zhang, ``Biometric
  recognition using deep learning: A survey,'' \emph{arXiv preprint
  arXiv:1912.00271}, 2019.

\bibitem{chen2017deeplab}
L.-C. Chen, G.~Papandreou, I.~Kokkinos, K.~Murphy, and A.~L. Yuille, ``Deeplab:
  Semantic image segmentation with deep convolutional nets, atrous convolution,
  and fully connected crfs,'' \emph{IEEE transactions on pattern analysis and
  machine intelligence}, vol.~40, no.~4, pp. 834--848, 2017.

\bibitem{minaee2019mtbi}
S.~Minaee, Y.~Wang, A.~Aygar, S.~Chung, X.~Wang, Y.~W. Lui, E.~Fieremans,
  S.~Flanagan, and J.~Rath, ``Mtbi identification from diffusion mr images
  using bag of adversarial visual features,'' \emph{IEEE transactions on
  medical imaging}, vol.~38, no.~11, pp. 2545--2555, 2019.

\bibitem{zhao2009direct}
Q.~Zhao, L.~Zhang, D.~Zhang, and N.~Luo, ``Direct pore matching for fingerprint
  recognition,'' in \emph{International Conference on Biometrics}.\hskip 1em
  plus 0.5em minus 0.4em\relax Springer, 2009, pp. 597--606.

\bibitem{minaee2015fingerprint}
S.~Minaee and Y.~Wang, ``Fingerprint recognition using translation invariant
  scattering network,'' in \emph{2015 IEEE Signal Processing in Medicine and
  Biology Symposium (SPMB)}.\hskip 1em plus 0.5em minus 0.4em\relax IEEE, 2015,
  pp. 1--6.

\bibitem{de2016iris}
M.~De~Marsico, A.~Petrosino, and S.~Ricciardi, ``Iris recognition through
  machine learning techniques: A survey,'' \emph{Pattern Recognition Letters},
  vol.~82, pp. 106--115, 2016.

\bibitem{minaee2016experimental}
S.~Minaee, A.~Abdolrashidiy, and Y.~Wang, ``An experimental study of deep
  convolutional features for iris recognition,'' in \emph{2016 IEEE signal
  processing in medicine and biology symposium (SPMB)}.\hskip 1em plus 0.5em
  minus 0.4em\relax IEEE, 2016, pp. 1--6.

\bibitem{minaee2015highly}
S.~Minaee and A.~Abdolrashidi, ``Highly accurate palmprint recognition using
  statistical and wavelet features,'' in \emph{2015 IEEE Signal Processing and
  Signal Processing Education Workshop (SP/SPE)}.\hskip 1em plus 0.5em minus
  0.4em\relax IEEE, 2015, pp. 31--36.

\bibitem{mistani2011multispectral}
S.~A. Mistani, S.~Minaee, and E.~Fatemizadeh, ``Multispectral palmprint
  recognition using a hybrid feature,'' \emph{arXiv preprint arXiv:1112.5997},
  2011.

\bibitem{wright2008robust}
J.~Wright, A.~Y. Yang, A.~Ganesh, S.~S. Sastry, and Y.~Ma, ``Robust face
  recognition via sparse representation,'' \emph{IEEE transactions on pattern
  analysis and machine intelligence}, vol.~31, no.~2, pp. 210--227, 2008.

\bibitem{parkhi2015deep}
O.~M. Parkhi, A.~Vedaldi, and A.~Zisserman, ``Deep face recognition,'' 2015.

\bibitem{minaee2017face}
S.~Minaee, A.~Abdolrashidi, and Y.~Wang, ``Face recognition using scattering
  convolutional network,'' in \emph{2017 IEEE Signal Processing in Medicine and
  Biology Symposium (SPMB)}.\hskip 1em plus 0.5em minus 0.4em\relax IEEE, 2017,
  pp. 1--6.

\bibitem{levi2015age}
G.~Levi and T.~Hassner, ``Age and gender classification using convolutional
  neural networks,'' in \emph{Proceedings of the IEEE conference on computer
  vision and pattern recognition workshops}, 2015, pp. 34--42.

\bibitem{duan2018hybrid}
M.~Duan, K.~Li, C.~Yang, and K.~Li, ``A hybrid deep learning cnn--elm for age
  and gender classification,'' \emph{Neurocomputing}, vol. 275, pp. 448--461,
  2018.

\bibitem{ozbulak2016transferable}
G.~Ozbulak, Y.~Aytar, and H.~K. Ekenel, ``How transferable are cnn-based
  features for age and gender classification?'' in \emph{2016 International
  Conference of the Biometrics Special Interest Group (BIOSIG)}.\hskip 1em plus
  0.5em minus 0.4em\relax IEEE, 2016, pp. 1--6.

\bibitem{lapuschkin2017understanding}
S.~Lapuschkin, A.~Binder, K.-R. Muller, and W.~Samek, ``Understanding and
  comparing deep neural networks for age and gender classification,'' in
  \emph{Proceedings of the IEEE International Conference on Computer Vision
  Workshops}, 2017, pp. 1629--1638.

\bibitem{rodriguez2017age}
P.~Rodr{\'\i}guez, G.~Cucurull, J.~M. Gonfaus, F.~X. Roca, and J.~Gonz{\`a}lez,
  ``Age and gender recognition in the wild with deep attention,'' \emph{Pattern
  Recognition}, vol.~72, pp. 563--571, 2017.

\bibitem{lee2018adversarial}
S.~S. Lee, H.~G. Kim, K.~Kim, and Y.~M. Ro, ``Adversarial spatial frequency
  domain critic learning for age and gender classification,'' in \emph{2018
  25th IEEE International Conference on Image Processing (ICIP)}.\hskip 1em
  plus 0.5em minus 0.4em\relax IEEE, 2018, pp. 2032--2036.

\bibitem{kim2018age}
H.~Kim, S.-H. Lee, M.-K. Sohn, and B.~Hwang, ``Age and gender estimation using
  region-sift and multi-layered svm,'' in \emph{Tenth International Conference
  on Machine Vision (ICMV 2017)}, vol. 10696.\hskip 1em plus 0.5em minus
  0.4em\relax International Society for Optics and Photonics, 2018, p. 106962J.

\bibitem{zhang2018landmark}
Y.~Zhang and T.~Xu, ``Landmark-guided local deep neural networks for age and
  gender classification,'' \emph{Journal of Sensors}, vol. 2018, 2018.

\bibitem{wang2017residual}
F.~Wang, M.~Jiang, C.~Qian, S.~Yang, C.~Li, H.~Zhang, X.~Wang, and X.~Tang,
  ``Residual attention network for image classification,'' in \emph{Proceedings
  of the IEEE conference on computer vision and pattern recognition}, 2017, pp.
  3156--3164.

\bibitem{he2016deep}
K.~He, X.~Zhang, S.~Ren, and J.~Sun, ``Deep residual learning for image
  recognition,'' in \emph{Proceedings of the IEEE conference on computer vision
  and pattern recognition}, 2016, pp. 770--778.

\bibitem{utk}
Z.~Zhang, Y.~Song, and H.~Qi, ``Age progression/regression by conditional
  adversarial autoencoder,'' in \emph{Proceedings of the IEEE conference on
  computer vision and pattern recognition}, 2017, pp. 5810--5818.

\bibitem{paszke2017automatic}
A.~Paszke, S.~Gross, S.~Chintala, G.~Chanan, E.~Yang, Z.~DeVito, Z.~Lin,
  A.~Desmaison, L.~Antiga, and A.~Lerer, ``Automatic differentiation in
  pytorch,'' 2017.

\bibitem{nair2019automated}
R.~R. Nair, R.~Madhavankutty, and S.~Nema, ``Automated detection of gender from
  face images,'' 2019.

\end{thebibliography}

%\bibliographystyle{unsrt} 
%\input{output.bbl}

\end{document}